\documentclass[10pt,letterpaper,twocolumn]{article}
\usepackage[T1]{fontenc}
\usepackage{newtxtext}
\usepackage{helvet}
\usepackage{courier}
\usepackage[margin=0.75in,columnsep=0.375in]{geometry}
\usepackage[hyphens]{url}
\usepackage{graphicx}
\usepackage[round,authoryear]{natbib}
\usepackage{caption}
\usepackage{booktabs}
\usepackage{amsmath}
\usepackage{amssymb}
\usepackage{tikz}
\usetikzlibrary{arrows.meta,decorations.pathreplacing}
\usepackage{placeins}
\usepackage[hidelinks,unicode]{hyperref}
\input{glyphtounicode}
\hypersetup{
  pdftitle={ActMap: Single-Pass Uncertainty Quantification from Generation-Time Activation Maps},
  pdfauthor={Jacopo Dardini; Roberta Calegari},
  pdfsubject={Uncertainty quantification for large language models},
  pdfkeywords={uncertainty quantification, large language models, activation maps}
}
\newcommand{\actmap}{\textsc{ActMap}}
\newcommand{\shapeactmap}{\(12 \times 32 \times 128\)}
\newcommand{\actvitqwenrow}{%
 & ACT-ViT & .861 & .856 & .059 & \textbf{.855} & \textbf{.827} & .088 & \textbf{.835} & \textbf{.844} & \textbf{.100} & \textbf{.717} & \textbf{.703} & .083 \\
}
\newcommand{\actvitllamarow}{%
 & ACT-ViT & .832 & .833 & .072 & .848 & .823 & .098 & .912 & .906 & \textbf{.061} & .766 & .753 & .103 \\
}
\newcommand{\actvitmistralrow}{%
 & ACT-ViT & \textbf{.885} & \textbf{.881} & .066 & \textbf{.911} & \textbf{.878} & .124 & .769 & .763 & .159 & \textbf{.689} & .664 & .077 \\
}

\title{\actmap{}: Single-Pass Uncertainty Quantification\\from Generation-Time Activation Maps}
\author{Jacopo Dardini\qquad Roberta Calegari\\[0.4em]
  \normalsize University of Bologna, Italy}
\date{}
\begin{document}

\maketitle

\begin{abstract}
Practical uncertainty quantification (UQ) for large language models must decide, from a single generation, whether a specific answer should be trusted. Existing methods either sample multiple generations, read only output-token probabilities, or reduce the model's internal computation to a single hidden state. We introduce \actmap{}, a white-box representation that compresses the generation-time hidden-state trajectory (every layer, every generated token) into a fixed \shapeactmap{} tensor of temporal-statistic channels that preserves structure across transformer depth and pooled hidden coordinates. The map is captured during the generation pass with no measurable overhead, has a fixed shape across model depths and hidden sizes, and occupies 96\,KiB: a compact artifact that can be retained for audit-relevant generations and probed directly, with occlusion analysis localizing the classifier's signal to mid-depth regions of the map. A lightweight classifier, instantiated as a compact Vision Transformer, reads an estimated correctness probability from each map in a fraction of a millisecond; capacity-matched MLPs perform comparably, indicating the representation itself carries the result. Trained and evaluated in-domain on short-answer QA, direct-answer math, and summarization factuality with three instruction-tuned 7--8B models, \actmap{} consistently outperforms sampling, token-probability, attention, and embedding baselines, and matches ACT-ViT, a detector trained on dense activation tensors 67$\times$ larger, at essentially the same mean AUROC with lower calibration error on ten of twelve pairs. The resulting score supports abstention, routing, and selective verification from a single generation, making it a practical primitive for scalable oversight of deployed models.
\end{abstract}

\section{Introduction}

Large language models answer questions, solve problems, and summarize documents with fluent confidence whether or not they are right. Deployments that act on these outputs need a per-answer reliability signal: a score that says, for this specific generation, how likely it is to be correct. This answer-level uncertainty quantification (UQ) problem, detecting hallucinated or otherwise unreliable answers, is increasingly the gatekeeper for abstention, retrieval fallback, escalation, and human review \citep{kadavath2022,farquhar2024}, and the signal that decides where scarce verification effort goes when human oversight cannot scale to every generation \citep{bowman2022}.

Existing UQ methods occupy three regimes, each with a structural gap. Sampling-based black-box methods such as semantic entropy measure disagreement across multiple generations \citep{kuhn2023,farquhar2024}; they capture meaning-level ambiguity but require $N$ sampled generations per query, which raises total decoding compute and throughput requirements even when parallel sampling hides latency. Grey-box methods read output-token probabilities (perplexity, mean token entropy, or learned functions of the output distribution \citep{barshalom2026los}) from a single pass, but see only the final projection of the model's computation. White-box methods look inside the model, yet most reduce the internal state to a single vector: a probe on the last token's hidden state \citep{azaria2023,marks2023}, a pooled sentence representation, or aggregated logits at one self-evaluation position \citep{xiao2026eagle}. Between ``one vector'' and ``ten regenerations'' lies almost everything the model computed while producing the answer: how representations differ across depth and across the hidden space, and how they evolve over generated tokens; many existing methods collapse or omit much of this structure.

We propose \actmap{}, a representation designed to keep it. During a single generation, \actmap{} records the hidden state of every transformer layer at every generated token, then compresses this variable-size $L \times T \times D$ trajectory into a fixed \shapeactmap{} tensor: twelve channels of temporal statistics over the token axis, with adaptive pooling mapping the layer and hidden-dimension axes to fixed sizes. The result is compact (96\,KiB per generation), fixed in shape across model depths and hidden sizes, and structured: rows index transformer depth, columns index pooled hidden coordinates, and channels are aligned temporal statistics. A learned classifier, in our main experiments a compact Vision Transformer, maps each tensor to an estimated correctness probability. The classifier never sees the generated text or output token probabilities: correctness is predicted from the internal state alone, with no extra model calls.

Our primary setting is in-domain deployment: an operator serves a fixed model on a fixed task, labels a set of generations once, and then assigns every subsequent answer a single correctness score, on which downstream decisions such as abstention, escalation, routing, or selective verification can be thresholded. Our main contributions are as follows:
\begin{itemize}
    \item \textbf{Representation.} A fixed-size, multi-channel activation-map summary of the full generation trajectory (layers, generated tokens, hidden dimensions, and activation dynamics), computable during one generation pass for any decoder-only transformer; ablations show robustness to the classifier choice.
    \item \textbf{Method.} A single-pass supervised UQ method with negligible added inference cost: capture is not measurably slower than plain decoding, and scoring is one forward pass of a 2.4M-parameter classifier.
    \item \textbf{Evaluation.} A unified comparison against eight baselines organized in an explicit black-/grey-/white-box taxonomy, on four tasks and three open-weight 7--8B models under a shared balanced protocol. The comparison includes ACT-ViT, evaluated with its complete published per-pair architecture sweep; \actmap{} reaches essentially the same mean AUROC from a 67$\times$ smaller representation with a single fixed classifier configuration.
    \item \textbf{Analysis.} Ablations and controls that identify cross-layer, pooled-coordinate structure as the primary source of predictive signal; an occlusion analysis of where the classifier draws that signal; transfer experiments across datasets, tasks, generators, and model scale (with clearly reported near-chance results); and calibration, selective-prediction, and cost analyses.
\end{itemize}

\section{Related Work}

\paragraph{Sampling-based black-box UQ.}
Semantic entropy samples several answers to the same query, clusters them by bidirectional entailment, and computes entropy over the resulting meaning classes \citep{kuhn2023,farquhar2024}. It needs no access to model internals, but each scored query requires $N$ sampled generations (here $N{=}10$) plus entailment inference.

\paragraph{Token-probability (grey-box) methods.}
Sequence perplexity and mean token entropy summarize the output distribution of a single generation; P(True)-style self-evaluation elicits the probability the model assigns to its own answer being correct \citep{kadavath2022}. LOS-Net learns a detector over the full sequence of next-token distributions \citep{barshalom2026los}, evidence that learned functions of output distributions can outperform hand-crafted statistics. All of these observe only the model's final projection onto the vocabulary.

\paragraph{Single-vector white-box probes.}
Linear probes on one hidden state can recover truthfulness information \citep{azaria2023,marks2023}, and layer-aggregated logits at a self-evaluation token improve calibration \citep{xiao2026eagle}. These approaches collapse the token axis entirely (one position) or the layer axis (one pooled vector), assuming the reliability signal is localized.

\paragraph{Structured white-box methods.}
EigenScore measures the differential entropy of sampled-response embeddings in internal space \citep{chen2024inside}; RAUQ aggregates attention through uncertainty-relevant heads, unsupervised, in a single pass \citep{vazhentsev2026rauq}; TAD learns attention-based features of conditional dependency between generation steps \citep{vazhentsev2024tad}. ACT-ViT is the closest representation-level comparison: it trains a vision transformer directly over padded layer-by-token activation tensors \citep{bar2025beyond}. Unlike methods that select attention statistics or embedding geometry, ACT-ViT and \actmap{} both preserve joint depth and token structure. They differ in what is retained: ACT-ViT keeps a dense tensor whose width tracks the generator's hidden size, reads at most the first 100 generated tokens, and couples the detector to the generator through a model-specific adapter; \actmap{} compresses the full trajectory into a fixed-shape map that is identical in geometry for every generator.

\paragraph{Summarization factuality.}
Summary factuality is judged against the source document and is known to be graded and multi-dimensional rather than binary \citep{maynez2020}. We use MiniCheck sentence-level verification for CNN/DailyMail \citep{tang2024minicheck} and assign one factual or non-factual label to each summary. This supports the same correctness-prediction interface for long-form generation, although summaries near the label boundary remain difficult.

\paragraph{Positioning.}
\actmap{} keeps what each family discards: it scores the single produced answer from one pass (vs.\ sampling), reads the pre-projection trajectory (vs.\ grey-box), preserves the depth axis and summarizes evolution over generated tokens (vs.\ single-vector probes), and exposes a broad structured summary for a learned classifier to mine rather than committing to one signal \citep{elhage2022}. Viewed as a monitor, \actmap{} extends work that reads safety-relevant signals directly from internal activations \citep{burns2023dlk}. Table~\ref{tab:taxonomy} places every baseline in this taxonomy.

\section{Method}

\subsection{Generation-Time Trajectory}

Let a decoder-only transformer with $L$ layers and hidden size $D$ generate an answer of $T$ tokens. During decoding we record, via forward hooks on every layer, the hidden state of each newly generated token, giving a trajectory
\begin{equation}
    H \in \mathbb{R}^{L \times T \times D},
\end{equation}
where $H_{\ell,t}$ is the layer-$\ell$ representation of generated token $t$. To bound memory, each hidden vector is reduced online from $D$ to $D'{=}128$ coordinates by contiguous adaptive average pooling (each pooled coordinate is the mean of a fixed contiguous block of hidden dimensions), so the stored trajectory is $L \times T \times 128$. Generation itself is unchanged: one decoding pass in vLLM \citep{kwon2023vllm}, no extra samples.

\begin{figure*}[t]
\centering
\begin{tikzpicture}[
  font=\small,
  stage/.style={draw, rounded corners=2.5pt, align=center, minimum height=1.15cm, inner sep=5pt},
  model/.style={stage, fill=orange!22, draw=orange!65!black},
  score/.style={stage, fill=green!16, draw=green!45!black},
  flow/.style={-{Stealth[length=2.4mm]}, semithick, draw=black!75},
  albl/.style={font=\scriptsize, align=center, text=black!75},
  nlbl/.style={font=\scriptsize, align=center}
]
\node[model] (llm) at (0.2,0) {LLM\\prompt $\to$ answer};
\begin{scope}[shift={(3.1,-0.575)}]
  \foreach \i in {2,1,0} {
    \draw[fill=blue!9, draw=blue!50!black, thin, shift={(0.11*\i,-0.11*\i)}] (0,0) rectangle (2.0,1.15);
    \draw[blue!30, very thin, shift={(0.11*\i,-0.11*\i)}, xstep=0.2, ystep=0.23] (0.001,0.001) grid (1.999,1.149);
  }
\end{scope}
\node[nlbl] at (4.2,-1.25) {hidden trajectory\\$L{\times}T{\times}D$};
\begin{scope}[shift={(8.55,0)}]
  \node[inner sep=1pt, fill=white, draw=black!50, very thin] at (0.28,-0.28) {\includegraphics[width=2.8cm]{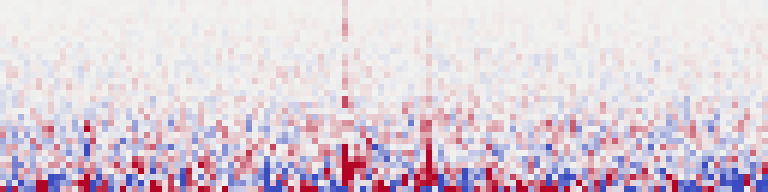}};
  \node[inner sep=1pt, fill=white, draw=black!50, very thin] at (0.14,-0.14) {\includegraphics[width=2.8cm]{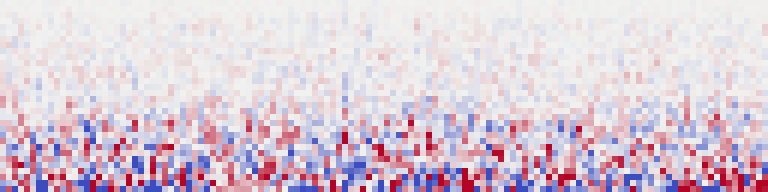}};
  \node[inner sep=1pt, fill=white, draw=black!65, very thin] (map) at (0,0) {\includegraphics[width=2.8cm]{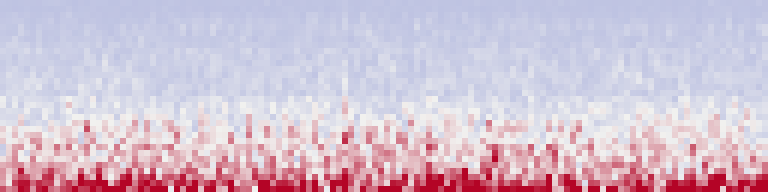}};
\end{scope}
\node[nlbl] at (8.55,-1.25) {\actmap{}\\$12{\times}32{\times}128$};
\node[model] (vit) at (13.0,0) {Vision Transformer\\classifier};
\node[score] (p) at (15.75,0) {$p(\text{correct})$};
\draw[flow] (llm.east) -- (3.0,0) node[albl, pos=0.5, above=2pt] {capture every\\layer \& token};
\draw[flow] (5.5,0) -- (7.0,0) node[albl, pos=0.55, above=2pt] {12 temporal stats,\\pool, z-score};
\draw[flow] (10.3,0) -- (vit.west) node[albl, pos=0.5, above=2pt] {single\\forward pass};
\draw[flow] (vit.east) -- (p.west) node[albl, midway, above=2pt] {$\sigma$};
\draw[decorate, decoration={brace, mirror, amplitude=4pt}, black!70]
  (-1.1,-1.75) -- (5.55,-1.75) node[albl, midway, below=5pt] {during generation (single pass)};
\draw[decorate, decoration={brace, mirror, amplitude=4pt}, black!70]
  (6.05,-1.75) -- (16.55,-1.75) node[albl, midway, below=5pt] {after generation (no extra LLM calls)};
\end{tikzpicture}
\caption{The \actmap{} pipeline. The LLM generates its answer normally while hooks capture the per-layer, per-token hidden-state trajectory; temporal statistics, pooling, and per-channel standardization compress it into a fixed \shapeactmap{} map, from which a compact Vision Transformer classifier reads $p(\text{correct})$. Channels shown: token standard deviation, temporal slope, first segment mean (real Qwen3-8B GSM8K generation).}
\label{fig:pipeline}
\end{figure*}
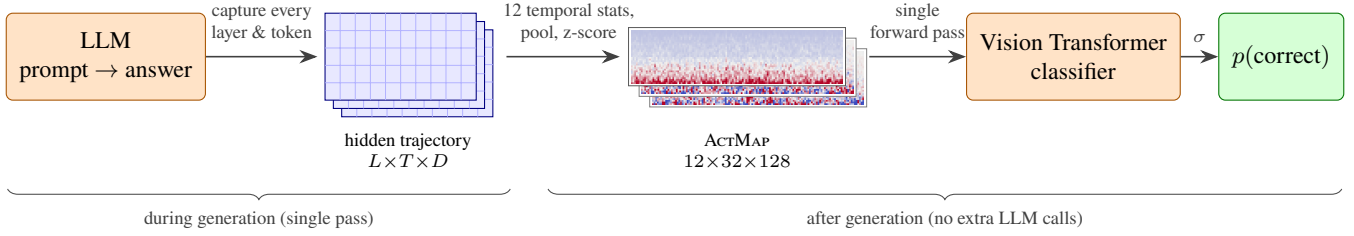

\subsection{Activation-Map Construction}

The trajectory varies in $T$ (answers have different lengths) and, across models, in $L$ and $D$. \actmap{} converts it into a fixed tensor $M \in \mathbb{R}^{C \times L' \times D'}$ with $C{=}12$, $L'{=}32$, $D'{=}128$ in three steps (Figure~\ref{fig:pipeline}).

\emph{(1) Temporal statistics.} The token axis is summarized by twelve channels, each a function $\mathbb{R}^{T} \to \mathbb{R}$ applied independently at every (layer, pooled-coordinate) location: \textbf{segment means} (4: means over the four consecutive quarters of the answer, locating activation mass in answer time), \textbf{final states} (2: last-token state and mean over the final eight tokens, the information a last-token probe would see), \textbf{dispersion} (2: standard deviation and maximum over tokens), \textbf{drift and slope} (2: last-minus-first difference and least-squares slope against token index), and \textbf{magnitude and dynamics} (2: per-layer RMS norm, broadcast over coordinates, and mean absolute token-to-token difference). Every channel is a statistic over the whole token axis, so the output is independent of $T$ by construction; one-token answers set the dispersion, slope, and dynamics channels to zero.

\emph{(2) Layer pooling.} The layer axis is adaptively average-pooled to $L'{=}32$ rows, aligning models with different depths ($L \in \{32, 36, 64\}$ here) onto a common axis while preserving depth ordering.

\emph{(3) Normalization.} Each channel is standardized to zero mean and unit variance over its $32 \times 128$ entries, so the classifier sees spatial \emph{patterns} within a channel rather than raw magnitudes, and channels are on a common scale. Maps are stored in float16 (96\,KiB per generation).

We deliberately avoid interpreting individual cells; the representation's role is to preserve \emph{where in depth}, \emph{where in the pooled hidden space}, and \emph{when in answer time} activity differs between correct and incorrect generations.

\subsection{Correctness Classifier}

The detector $f_\theta$ maps a tensor $M$ to an estimated correctness probability $p(\text{correct}) = \sigma(f_\theta(M))$; the uncertainty score of a generation is $u = 1 - p(\text{correct})$. We instantiate $f_\theta$ as a compact Vision Transformer \citep{dosovitskiy2021} over $4 \times 16$ patches of the map (2.4M parameters; six pre-norm blocks, embedding width 192, six heads, a class token) with factorized row/column positional embeddings, so that depth and coordinate identity are preserved; remaining details appear in the appendix. The representation is not tied to this choice (see Ablations); all main-table results use the Vision Transformer.

Training minimizes binary cross-entropy on maps with binary correctness labels, using AdamW (learning rate $10^{-3}$, weight decay 0.05), cosine decay with 5 warm-up epochs, at most 80 epochs with early stopping on validation AUROC (patience 20), batch size 64, Gaussian input noise ($\sigma{=}0.08$) and mixup ($\alpha{=}0.2$), and three seeds $\{42, 123, 456\}$. Reported probabilities are raw sigmoid outputs; deployment at a base rate different from the training distribution may require prior correction (see Calibration). Scoring a stored map is a single forward pass of the small network; no additional LLM call is made.

\section{Experimental Setup}

\subsection{Tasks, Models, and Generation}

We evaluate on four datasets covering three task families and two output-length regimes (Table~\ref{tab:datasets}): TriviaQA (no context) and NQ-Open for short-answer factual QA \citep{triviaqa,naturalquestions}, GSM8K for mathematical problem solving in direct-answer mode \citep{gsm8k}, and CNN/DailyMail for long-form summarization factuality \citep{cnn_dailymail}. Generators are Qwen3-8B \citep{qwen3}, Llama-3.1-8B-Instruct \citep{llama3}, and Mistral-7B-Instruct-v0.3 \citep{mistral7b}; Qwen3-32B (64 layers) is used for an in-domain and model-scale transfer study on the three short-form tasks. Primary answer generation is greedy (temperature 0) in vLLM with a 4{,}096-token context, 32 new tokens for short-answer tasks and GSM8K (direct answer, no explicit reasoning), and 384 for summaries. In total, the study produces more than 476{,}000 generations with captured trajectories.

\begin{table}[t]
\centering
\caption{Datasets, pre-balancing split sizes, and label types.}
\label{tab:datasets}
\small
\setlength{\tabcolsep}{3.5pt}
\begin{tabular}{lrrl}
\toprule
Dataset & Rows & Test & Label \\
\midrule
TriviaQA (no ctx.) & 50{,}000 & 5{,}051 & answer correctness \\
NQ-Open & 50{,}000 & 5{,}064 & answer correctness \\
GSM8K (direct) & 8{,}792 & 879 & numeric correctness \\
CNN/DailyMail & 50{,}000 & 5{,}000 & summary factuality \\
\bottomrule
\end{tabular}
\end{table}

\subsection{Labels and Splits}

For TriviaQA and NQ-Open, a response is correct if its normalized answer exactly matches a normalized gold alias or attains token F1 of at least 0.8 against any alias; GSM8K uses exact numeric match. CNN/DailyMail summaries are split into sentences and each sentence is verified against the source article with MiniCheck (Flan-T5-Large); a summary is labeled factual iff every sentence is supported at threshold 0.5. Splits are disjoint by source key. Because per-model accuracy varies, all supervised training and all reported metrics use per-(dataset, model, split) \emph{balanced} indices with equal numbers of correct and incorrect generations. All supervised detectors (\actmap{}, TAD, and ACT-ViT) train on the same balanced training rows, use the same balanced validation rows for their respective model-selection protocols, and are evaluated on the identical balanced test rows.

\subsection{Baseline Taxonomy}

Table~\ref{tab:taxonomy} summarizes all methods. The black-box baseline is Semantic Entropy (10 sampled generations at temperature 1.0, top-$p$ 0.9, clustered by bidirectional DeBERTa-MNLI entailment). Grey-box baselines are sequence perplexity, mean token entropy (MTE), and P(True) self-evaluation. White-box baselines are TAD and ACT-ViT (supervised, like \actmap{}), RAUQ, and EigenScore (over 10 sampled-response embeddings). We follow each baseline's published protocol; for ACT-ViT this includes the authors' full 24-configuration architecture sweep on the shared balanced splits and seeds (details in the appendix).

\begin{table*}[t]
\centering
\caption{Method taxonomy. ``Extra gen.'' counts generated responses beyond the scored answer. Per-answer latencies, which are specific to our implementation and hardware stack, are reported in the appendix.}
\label{tab:taxonomy}
\scriptsize
\setlength{\tabcolsep}{3.5pt}
\begin{tabular}{llclcl}
\toprule
Method & Access & Extra gen. & Information used & Supervision & Stored artifact \\
\midrule
Semantic Entropy \citep{kuhn2023} & black-box & 10 & sampled answer texts & none & samples \\
\midrule
Perplexity & grey-box & 0 & output token probabilities & none & -- \\
Mean token entropy & grey-box & 0 & output token distributions & none & -- \\
P(True) \citep{kadavath2022} & grey-box & 1 short pass & self-evaluation token probability & none & -- \\
\midrule
TAD \citep{vazhentsev2024tad} & white-box & 0 & attention + token probabilities & trained & attn.\ features \\
RAUQ \citep{vazhentsev2026rauq} & white-box & 0 & attention + token probabilities & none & attn.\ stats \\
EigenScore \citep{chen2024inside} & white-box & 10 & sampled-response embeddings & none & embeddings \\
ACT-ViT \citep{bar2025beyond} & white-box & 0 & padded activation tensor & trained & activations \\
\actmap{} (ours) & white-box & 0 & full hidden-state trajectory & trained & 96\,KiB map \\
\bottomrule
\end{tabular}
\end{table*}

\subsection{Metrics and Evaluation Axes}

We report AUROC ($\uparrow$; ranking quality across all thresholds), AUPRC ($\uparrow$; precision--recall performance, with a 0.5 baseline on the balanced splits), and 10-bin expected calibration error (ECE, $\downarrow$; the gap between predicted confidence and empirical correctness) \citep{guo2017}. Supervised detectors report the mean over seeds $\{42,123,456\}$; the maximum seed standard deviation of \actmap{} AUROC is 0.026 (GSM8K, smallest split) and below 0.010 on all splits above 1{,}700 test rows. The evaluation separates (i) \emph{in-domain} performance, training and testing on the same (dataset, model); (ii) \emph{cross-dataset} transfer within a model; (iii) \emph{cross-task} transfer between short-form correctness and long-form factuality; (iv) \emph{cross-generator} transfer; and (v) \emph{model-scale} transfer (Qwen3-8B$\leftrightarrow$32B).

\section{Main Results}

Table~\ref{tab:main} reports the in-domain matrix over all (model, dataset, method) triplets.

\begin{table*}[t]
\centering
\caption{In-domain results on balanced test splits (AUROC $\uparrow$ / AUPRC $\uparrow$ / ECE $\downarrow$; mean over three seeds for trained methods). \textbf{Bold} marks the best value for that pair. ECE for methods without native probabilities uses a fixed monotone score-to-probability mapping. CNN/DailyMail EigenScore covers the label subset with sampled-response embeddings.}
\label{tab:main}
\scriptsize
\setlength{\tabcolsep}{1.9pt}
\renewcommand{\arraystretch}{1.0}
\begin{tabular}{llcccccccccccc}
\toprule
Model & Method & \multicolumn{3}{c}{TriviaQA} & \multicolumn{3}{c}{NQ-Open} & \multicolumn{3}{c}{GSM8K} & \multicolumn{3}{c}{CNN/DM} \\
\cmidrule(lr){3-5}\cmidrule(lr){6-8}\cmidrule(lr){9-11}\cmidrule(lr){12-14}
 & & AUROC & AUPRC & ECE & AUROC & AUPRC & ECE & AUROC & AUPRC & ECE & AUROC & AUPRC & ECE \\
\midrule
Qwen3-8B
 & \actmap{} (ours) & \textbf{.887} & \textbf{.883} & \textbf{.028} & .838 & .802 & \textbf{.057} & .791 & .781 & .126 & .705 & .692 & .033 \\
\actvitqwenrow
 & Semantic Entropy & .772 & .678 & .397 & .752 & .682 & .382 & .719 & .651 & .359 & .510 & .516 & .367 \\
 & Perplexity & .837 & .836 & .116 & .761 & .753 & .079 & .744 & .753 & .134 & .579 & .569 & \textbf{.028} \\
 & Mean token entropy & .847 & .841 & .186 & .760 & .751 & .116 & .755 & .758 & .142 & .582 & .570 & .061 \\
 & P(True) & .817 & .773 & .367 & .729 & .691 & .396 & .685 & .704 & .364 & .534 & .506 & .497 \\
 & TAD & .837 & .814 & .126 & .823 & .782 & .305 & .761 & .734 & .118 & .684 & .668 & .058 \\
 & RAUQ & .813 & .771 & .329 & .792 & .740 & .335 & .750 & .692 & .335 & .583 & .579 & .357 \\
 & EigenScore & .742 & .694 & .124 & .730 & .674 & .140 & .714 & .718 & .157 & .573 & .555 & .469 \\
\addlinespace[2pt]
Llama-3.1-8B
 & \actmap{} (ours) & \textbf{.880} & \textbf{.878} & \textbf{.039} & \textbf{.861} & \textbf{.840} & \textbf{.066} & \textbf{.931} & \textbf{.939} & .086 & \textbf{.769} & \textbf{.754} & \textbf{.044} \\
\actvitllamarow
 & Semantic Entropy & .802 & .744 & .386 & .758 & .719 & .368 & .575 & .582 & .228 & .483 & .488 & .418 \\
 & Perplexity & .830 & .834 & .169 & .795 & .793 & .150 & .812 & .772 & .177 & .576 & .560 & .054 \\
 & Mean token entropy & .844 & .843 & .171 & .796 & .792 & .155 & .827 & .793 & .233 & .579 & .559 & .111 \\
 & P(True) & .792 & .780 & .316 & .707 & .707 & .274 & .844 & .867 & .079 & .557 & .542 & .405 \\
 & TAD & .849 & .839 & .099 & .824 & .798 & .214 & .845 & .819 & .076 & .732 & .716 & .079 \\
 & RAUQ & .827 & .799 & .350 & .817 & .782 & .355 & .770 & .684 & .383 & .580 & .565 & .366 \\
 & EigenScore & .713 & .692 & .125 & .782 & .754 & .181 & .535 & .527 & .298 & .606 & .587 & .384 \\
\addlinespace[2pt]
Mistral-7B
 & \actmap{} (ours) & .859 & .851 & \textbf{.048} & .905 & .877 & \textbf{.041} & \textbf{.782} & \textbf{.813} & .146 & \textbf{.689} & \textbf{.671} & \textbf{.039} \\
\actvitmistralrow
 & Semantic Entropy & .711 & .614 & .445 & .777 & .699 & .410 & .728 & .673 & .288 & .495 & .510 & .413 \\
 & Perplexity & .777 & .763 & .070 & .853 & .857 & .050 & .751 & .745 & .142 & .516 & .514 & .041 \\
 & Mean token entropy & .787 & .767 & .144 & .850 & .854 & .172 & .750 & .745 & \textbf{.131} & .517 & .513 & .089 \\
 & P(True) & .774 & .730 & .390 & .731 & .663 & .426 & .575 & .622 & .504 & .577 & .556 & .500 \\
 & TAD & .788 & .748 & .141 & .891 & .856 & .337 & .771 & .768 & .153 & .676 & .661 & .082 \\
 & RAUQ & .761 & .724 & .329 & .879 & .858 & .341 & .757 & .759 & .343 & .521 & .518 & .363 \\
 & EigenScore & .681 & .676 & .418 & .855 & .828 & .298 & .597 & .585 & .433 & .556 & .541 & .208 \\
\bottomrule
\end{tabular}
\end{table*}

\paragraph{\actmap{} leads every non-tensor baseline.}
\actmap{} attains the highest AUROC and AUPRC on all twelve (model, dataset) pairs against the sampling, token-probability, attention, and embedding baselines. Its mean AUROC over those twelve pairs is .825, against .790 for TAD and .741 for the best training-free baseline (MTE).

\paragraph{Baseline categories behave consistently.}
Grey-box statistics are strong on short-answer QA, where a wrong answer usually coincides with a diffuse output distribution, but weaker on direct-answer mathematical correctness, where an incorrect numeric answer may still be produced with a concentrated token distribution: the best grey-box AUROC on GSM8K trails \actmap{} by .031--.087. Sampling-based Semantic Entropy pays for its ten extra generations without matching single-pass grey-box statistics on these balanced splits, consistent with its score reflecting question ambiguity rather than answer-specific reliability. Among the non-tensor white-box baselines, supervised TAD is consistently the strongest; unsupervised RAUQ and EigenScore sit between grey-box statistics and the supervised methods.

\paragraph{Long-form factuality is the hardest regime.}
On CNN/DailyMail, absolute scores drop for all methods. Summary factuality is graded: a mostly supported summary may contain one unsupported clause, making factual/non-factual labels difficult near the boundary. Training-free methods are barely better than chance here, and Semantic Entropy is at chance, since whole-response equivalence clustering is poorly matched to long-form output; only the supervised methods extract usable signal, with \actmap{} ahead of TAD on all three models, within .012 AUROC of ACT-ViT on every model, and keeping ECE below .05 throughout.

\subsection{Compression Preserves the Dense-Tensor Signal}

We evaluate ACT-ViT with its complete published 24-configuration sweep per pair rather than as a fixed-score baseline. Ranking quality is nearly identical: \actmap{} leads on seven of twelve pairs (one decided beyond three decimals) and ACT-ViT on five. Mean AUROC is .825 versus .823; the largest gaps are comparable (.048 and .044).

\actmap{} reaches this parity from 49{,}152 values per generation against the 3.3M values of ACT-ViT's dense $8{\times}100{\times}4096$ tensor, with one fixed classifier for all pairs while the sweep selects different architectures for different pairs, and with lower ECE on ten of twelve pairs (mean .063 vs.\ .091). These results indicate that temporal-statistic compression preserves the uncertainty signal in the dense activation tensor while producing a fixed, generator-agnostic map.

\section{Transfer and Generalization}

A detector is deployed either \emph{in-domain}, trained for a fixed generator and task (the previous section), or in \emph{transfer}, where target labels are unavailable and a detector trained elsewhere must generalize. We quantify the latter along four axes with the classifier frozen after source training (Table~\ref{tab:transfer}).

\begin{table}[t]
\centering
\caption{Frozen-detector transfer (AUROC $\uparrow$ / AUPRC $\uparrow$ / ECE $\downarrow$ on balanced target test splits; no target-domain training). Rows are macro means over their constituent target pairs and three seeds; model-scale and in-domain-scale rows average over the three short-form datasets.}
\label{tab:transfer}
\scriptsize
\setlength{\tabcolsep}{1.6pt}
\begin{tabular}{llccc}
\toprule
Axis & Source $\rightarrow$ target & AUROC & AUPRC & ECE \\
\midrule
Cross-dataset & TriviaQA $\rightarrow$ NQ-Open (3 models) & .837 & .821 & .138 \\
 & \quad frozen TAD, same transfer & .804 & .766 & .053 \\
Cross-task & TriviaQA+NQ $\rightarrow$ GSM8K & .617 & .645 & .304 \\
 & TriviaQA+NQ $\rightarrow$ CNN/DM & .512 & .513 & .463 \\
 & CNN/DM $\rightarrow$ TriviaQA+NQ & .537 & .532 & .256 \\
Cross-generator & leave-one-generator-out (12 targets) & .510 & .516 & .337 \\
Model scale & Qwen3-8B $\rightarrow$ 32B & .520 & .518 & .366 \\
 & Qwen3-32B $\rightarrow$ 8B & .517 & .522 & .403 \\
Scale (in-dom.) & Qwen3-32B, short-form & .868 & .860 & .065 \\
\bottomrule
\end{tabular}
\end{table}

Within-task transfer is strong: TriviaQA $\rightarrow$ NQ-Open stays close to matched in-domain training and ahead of the frozen TAD baseline, though TAD transfers with better calibration. Beyond the task boundary the picture changes: transfer from pooled QA to GSM8K is weak and model-dependent, and both directions between short-answer correctness and summary factuality are near chance. Cross-generator transfer trains on the pooled maps of two generators with the dataset held fixed and evaluates on the held-out third (a macro mean over four datasets, three targets, and three seeds); it is near chance. Model scale behaves the same way: transfer between Qwen3-8B and Qwen3-32B is near chance in both directions despite the shared model family and tokenizer, while a detector trained \emph{in domain} on Qwen3-32B is at least as strong as at 8B (macro .868 AUROC, with the largest gain on GSM8K, .791 $\to$ .870). These results concern transfer of the learned decision boundary. Shared map geometry alone does not align boundaries across tasks, generators, or scales. A new deployment therefore requires target-domain labels; otherwise the monitor degrades silently under shift, which is itself an oversight risk.

\section{Ablations}

All ablations run on TriviaQA $\times$ Qwen3-8B with the main-table protocol and three seeds (Table~\ref{tab:ablations}); they answer four questions.

\begin{table}[!t]
\centering
\caption{Ablations and controls on TriviaQA $\times$ Qwen3-8B (test AUROC, mean over three seeds; seed std $\leq$ .010 for every variant). Variants requiring re-captured trajectories use the retained class-balanced split intersection, on which the full configuration scores .886 (vs.\ .887 in Table~\ref{tab:main}).}
\label{tab:ablations}
\scriptsize
\setlength{\tabcolsep}{4pt}
\renewcommand{\arraystretch}{1.0}
\begin{tabular}{lc}
\toprule
Configuration & AUROC \\
\midrule
Full \actmap{} (12 ch., $32{\times}128$, Vision Transformer) & .886 \\
\addlinespace[2pt]
\emph{Representation scope} & \\
\quad last-token channels only & .879 \\
\quad single mean-pooled final-layer state & .754 \\
\emph{Pooling structure (controls)} & \\
\quad hidden-coordinate permutation before pooling & .713 \\
\quad Gaussian projection instead of pooling & .863 \\
\emph{Construction choices} & \\
\quad any one channel group removed (worst--best of 5) & .881--.890 \\
\quad 1 / 8 temporal segments (default 4) & .889 / .889 \\
\quad resolution $12 \times 16 \times 64$ & .877 \\
\quad resolution $12 \times 64 \times 256$ & .886 \\
\quad global normalization (not per-channel) & .889 \\
\emph{Classifier on identical maps} & \\
\quad logistic regression (flattened) & .877 \\
\quad MLP (matched parameters) & .892 \\
\quad MLP (4$\times$ parameters) & .893 \\
\emph{Training-set size} & \\
\quad 10\% / 25\% / 50\% of train & .815 / .855 / .875 \\
\emph{Sanity controls} & \\
\quad permuted labels (expect $\approx$.5) & .510 \\
\quad answer-length-only predictor & .597 \\
\bottomrule
\end{tabular}
\end{table}

\paragraph{Where does the gain come from?}
From the structure preserved across depth and pooled hidden coordinates; no single statistic explains it. Collapsing the map to one mean-pooled final-layer vector, the representation prior white-box probes use, costs .13 AUROC. A map built from the last-token channels alone recovers nearly all of the full map's performance, as expected on short answers where the final state can summarize the preceding computation. Temporal summaries add little beyond the last-token channels in both output-length regimes (CNN/DailyMail $\times$ Qwen3-8B, summaries up to 384 tokens: .704 vs.\ .705), and no single channel group is critical. The temporal channels are thus a compact mechanism for reducing variable-length trajectories to a fixed shape; the predictive signal lies in the cross-layer, pooled-coordinate structure.

\paragraph{Does the pooling scheme matter?}
Yes. Permuting hidden coordinates before pooling preserves the marginal activation values but destroys the coordinate grouping; it is the most damaging representation variant, costing more than collapsing the map to a single final-layer vector. Random Gaussian projections of matched size also lose ground. Performance depends on the consistent coordinate grouping induced by contiguous pooling; dimension reduction alone does not preserve the signal.

\paragraph{Does performance depend on the classifier architecture?}
No: logistic regression, a capacity-matched MLP, and a $4\times$ MLP all perform comparably on identical maps, and the capacity-matched MLP in fact slightly outperforms the Vision Transformer; on this ablation setting, classifier choice has little effect relative to the representation. The remaining construction choices (segment count, map resolution, normalization) shift AUROC by at most .010.

\paragraph{How much supervision is needed?}
On TriviaQA $\times$ Qwen3-8B, 10\% of the training data (about 3{,}800 balanced examples; AUROC .815) beats Semantic Entropy and EigenScore; 25\% (.855) beats every evaluated training-free baseline. As expected, a detector fit to permuted labels falls to chance, and an answer-length-only predictor stays well below the full map.

\subsection{Depth-Wise Localization of Predictive Signal}

\begin{figure*}[t]
\centering
\includegraphics[width=\textwidth]{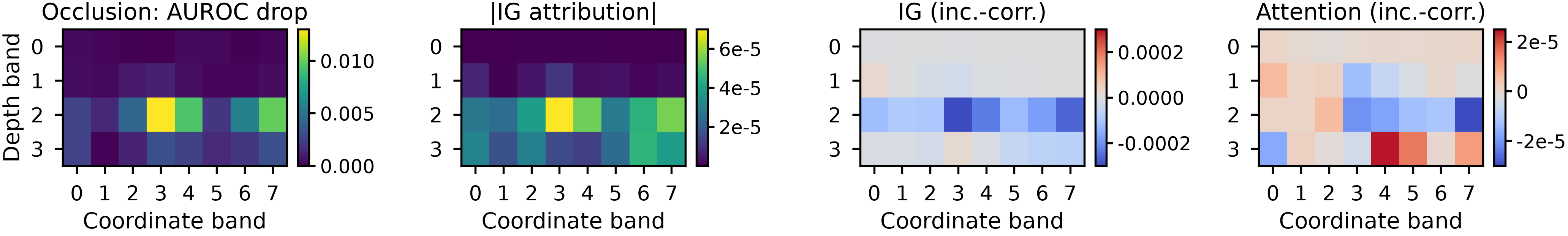}
\caption{Occlusion atlas for TriviaQA $\times$ Qwen3-8B (mean over three seeds): AUROC drop from zeroing one depth $\times$ coordinate band, with integrated-gradients (IG) and attention-rollout views of the same grid. All views concentrate in mid-depth bands.}
\label{fig:reliability-atlas}
\end{figure*}

We localize the predictive signal the classifier uses within the map: on TriviaQA $\times$ Qwen3-8B, we occlude one depth-band $\times$ coordinate-band region at a time and re-evaluate the frozen detector (baseline .887 AUROC; occluding the entire map collapses it to .500). The classifier is most sensitive to mid-network depth bands (occlusion drops of .003--.013), while the earliest and latest bands are individually more redundant; integrated-gradients and attention-rollout attributions agree on the same mid-depth concentration (Figure~\ref{fig:reliability-atlas}). This agreement localizes the detector's signal without implying an explicit correctness representation in the generator, and is consistent with probing literature placing semantic and truthfulness information in intermediate layers \citep{azaria2023,marks2023}.

\section{Calibration and Selective Prediction}

Abstention thresholds require calibrated probabilities, not only good ranking \citep{guo2017}. \actmap{}'s raw sigmoid outputs are the best- or near-best-calibrated on short-answer QA and CNN/DailyMail. The exception is GSM8K, whose training split is an order of magnitude smaller: \actmap{}'s ECE rises to .086--.146 and ACT-ViT or MTE is better calibrated there (Table~\ref{tab:main}). Raw-score ECE comparisons favor trained probabilistic detectors by construction. As a diagnostic, we fit one scalar temperature per method on half of the Qwen3-8B $\times$ TriviaQA test predictions and report ECE on the other half: \actmap{}'s optimal temperature is $\approx$1.0 (ECE .027 $\to$ .024), while MTE needs strong sharpening ($T{=}0.50$) and still leaves ECE at .158.

Balanced splits also differ from deployment prevalence, so we simulate prevalence shift by class-conditionally resampling each balanced test set to the model's natural test accuracy (100 replicates; temperature fit on 256 held-out target examples each). Where natural accuracy is near the balanced regime (TriviaQA, .53--.66; CNN/DailyMail, .54--.63), raw ECE stays at .02--.09 and selective prediction remains useful (18--38\% coverage at 5\% risk on TriviaQA). Where accuracy collapses (NQ-Open, .17--.23; Mistral-7B GSM8K, .07), ECE rises to .15--.24 and temperature scaling does not repair it: the error is a prior shift requiring prior correction, not a sharpness error.

\begin{figure}[t]
\centering
\includegraphics[width=\columnwidth]{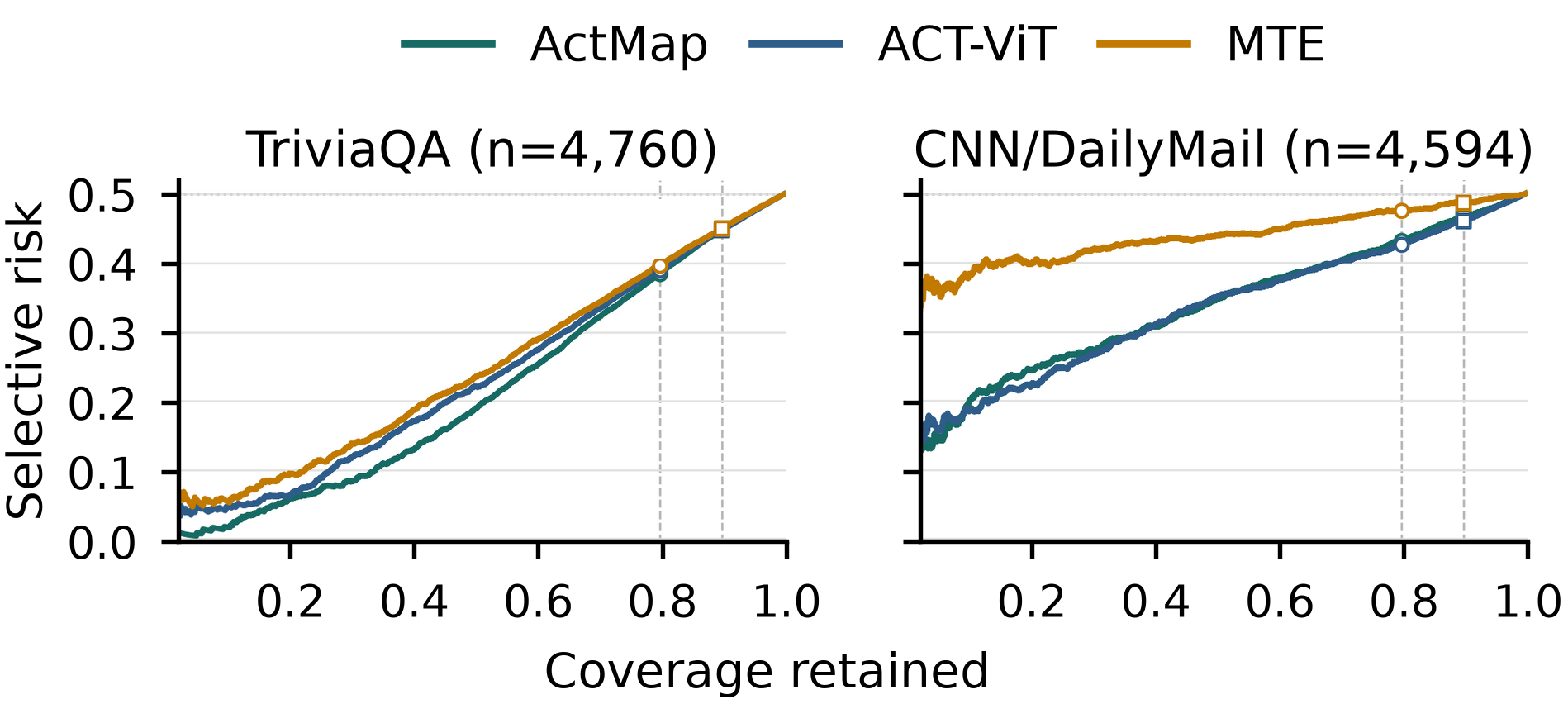}
\caption{Risk--coverage for Qwen3-8B: selective risk (error among retained answers) vs.\ coverage on shared balanced test rows. Markers: 80\% and 90\% coverage.}
\label{fig:risk-coverage}
\end{figure}

The deployment-relevant summary is risk--coverage behavior \citep{geifman2017}: how much generation volume can be cleared automatically at a target error rate, concentrating human review on the remainder (Figure~\ref{fig:risk-coverage}). On Qwen3-8B TriviaQA, \actmap{} retains 18.3\% coverage at 5\% risk versus 14.3\% for ACT-ViT and 6.2\% for MTE. CNN/DailyMail is substantially harder: no method provides useful coverage at 5\% risk; at 80\% coverage \actmap{} and ACT-ViT are comparable (.432 and .427) against .476 for MTE.

\section{Computational Cost}

Measurements use vLLM on one NVIDIA L40S with Qwen3-8B $\times$ TriviaQA; matched baseline latencies are in the appendix. \emph{Capture}: forward hooks pool every layer online; hooked and unhooked decoding both averaged about 7\,ms/prompt (five batches of 64), so overhead is within run-to-run variation. \emph{Storage}: one \shapeactmap{} float16 map is 96\,KiB (85--96$\times$ smaller than a 32-token float16 trajectory and 67$\times$\ smaller than ACT-ViT's dense tensor); 38k TriviaQA generations occupy 3.5\,GiB as maps versus 235\,GiB as dense tensors. \emph{Scoring}: classifier inference takes .024\,ms/map (batch 256), and training converges in under 20 GPU-minutes per seed. ACT-ViT takes .066\,ms/answer with 2.1$\times$ the parameters and 25$\times$ the peak memory; its CNN/DailyMail configuration trains for about 3 GPU-hours per pair before the 24-configuration sweep. Semantic Entropy and EigenScore require 10 additional generations per answer.

\section{Limitations and Ethical Considerations}

\paragraph{Limitations.}
\actmap{} requires white-box access to generation-time hidden states: self-hosted or provider-instrumented models, not closed APIs. It is supervised: each deployment regime needs its own labeled generations. Label sources are imperfect (alias matching misses paraphrases; MiniCheck inherits its judge's errors), and the score is correlational: high confidence means the trajectory resembles previously correct generations. It does not establish truth. Balanced-split evaluation differs from deployment prevalence (see Calibration). Our experiments use greedy decoding on English tasks; robustness across decoding strategies and temperatures is left to future work. In-domain results cover 7--8B generators on four tasks and Qwen3-32B on three short-form tasks; ablations cover two pairs.

\paragraph{Ethical considerations.}
A correctness score can reduce overreliance on fluent wrong answers, but a miscalibrated score can become a safety veneer. Under prevalence shift on NQ-Open and Mistral-7B GSM8K, ECE reaches .15--.24 and temperature scaling does not repair it. The score should guide verification rather than replace it. Stored maps require the source text's access controls and retention limits and provide a re-scorable audit trail for consequential answers.

\section{Conclusion}

\actmap{} converts generation-time hidden activations into a fixed-size representation over depth and pooled hidden coordinates. Trained in-domain, its lightweight classifier outperforms every evaluated non-tensor baseline on all twelve 7--8B pairs and matches dense activation-tensor learning from a 67$\times$ smaller, generator-agnostic artifact with better calibration. Capture adds no measurable overhead; results survive classifier swaps and hold in-domain at 32B scale. The main limitation is transfer: decision boundaries do not yet align across tasks, generators, or scales. Code, replication materials, and the 476{,}372-map dataset will be released on Hugging Face upon publication.

\FloatBarrier
\bibliographystyle{plainnat}
\bibliography{references}

\clearpage
\appendix
\setcounter{secnumdepth}{2}
\section*{Technical Appendix}
\phantomsection
\addcontentsline{toc}{section}{Technical Appendix}
This appendix gives the implementation details for the representation and
classifier, specifies the ACT-ViT reproduction protocol, and reports measured
costs. It also records the software environment and retained artifacts.
The main paper contains all claims and results needed to assess the paper.

\section{Exact Activation-Map Construction}

For one generated answer, let
$H\in\mathbb{R}^{L\times T\times D}$ contain the hidden state from every
transformer block and generated token. Hidden coordinates are first reduced to
$D'=128$ by one-dimensional adaptive average pooling over contiguous coordinate
intervals. Denote the resulting tensor by
$X\in\mathbb{R}^{L\times T\times D'}$.

The twelve channels below are computed independently for every layer $\ell$
and pooled coordinate $d$. For clarity, $x_t=X_{\ell,t,d}$ and
$\bar{x}=T^{-1}\sum_t x_t$.

\begin{enumerate}
    \item \textbf{Four segment means.} The token indices are divided by integer
    boundaries obtained from $\operatorname{linspace}(0,T,5)$. Each channel is
    the mean within one consecutive segment; an empty segment for a very short
    answer is replaced by its nearest valid one-token interval.
    \item \textbf{Final state.} $x_{T-1}$.
    \item \textbf{Final-window mean.}
    $\frac{1}{\min(8,T)}\sum_{t=\max(0,T-8)}^{T-1}x_t$.
    \item \textbf{Temporal standard deviation.} The sample standard deviation
    over tokens. It is zero for $T=1$.
    \item \textbf{Temporal maximum.} $\max_t x_t$.
    \item \textbf{Endpoint drift.} $x_{T-1}-x_0$.
    \item \textbf{Least-squares slope.}
    \begin{equation}
      \frac{\sum_t (t-\bar{t})(x_t-\bar{x})}
      {\max\{\sum_t(t-\bar{t})^2,10^{-8}\}},
    \end{equation}
    set to zero for $T=1$.
    \item \textbf{Layer RMS.} Unlike the other channels, this value is computed
    once per layer over all tokens and pooled coordinates:
    $\lVert X_{\ell,:,:}\rVert_2/\sqrt{TD'}$. It is then broadcast across the
    coordinate axis.
    \item \textbf{Mean absolute temporal difference.}
    $(T-1)^{-1}\sum_{t=1}^{T-1}|x_t-x_{t-1}|$, set to zero for $T=1$.
\end{enumerate}

Stacking the channels gives $S\in\mathbb{R}^{12\times L\times128}$.
Contiguous adaptive average pooling maps the ordered layer axis to 32 rows.
Finally, every channel is standardized independently within each example:
\begin{equation}
  M_c = \frac{S_c-\mu(S_c)}{\max\{\operatorname{std}(S_c),10^{-6}\}}.
\end{equation}
The stored map $M$ has shape $12\times32\times128$ and float16 size 96\,KiB.
All statistics are computed in float32 before storage.

\subsection{Design Rationale for the Twelve Channels}

\begin{table}[t]
\centering
\caption{Why each hand-designed channel family is included. These working
hypotheses are evaluated by the grouped ablations in the main paper; they do
not prescribe what the classifier must use.}
\label{tab:channel-rationale}
\small
\setlength{\tabcolsep}{4pt}
\begin{tabular}{p{0.27\linewidth}p{0.65\linewidth}}
\toprule
Channel family & Intended observable \\
\midrule
Segment means & Where activation mass lies during the answer and how its depth
profile changes over time. \\
Final state and window & The endpoint representation and its recent context,
comparable to last-token and pooled-state probes. \\
Dispersion & Variation within an answer and rare activation excursions that
may separate stable from conflicted or unstable generations. \\
Drift and slope & Whether activation patterns accumulate, resolve, or change
direction during the answer. \\
Layer RMS & Layer-wise activation energy independent of coordinate, retaining
a coarse depth profile. \\
Temporal differences & Local token-to-token changes that a global mean,
endpoint, or linear slope can miss. \\
\bottomrule
\end{tabular}
\end{table}

We chose a small, hand-designed channel set and fixed it before the final
benchmark evaluation. We do not claim that this set is unique or optimal. The
channels summarize four aspects of a generated answer: when a pattern occurs,
endpoint state, global dispersion, and temporal change. Segment means and
endpoint channels retain coarse phase information. Standard deviation and maximum describe the
distribution. Drift and slope capture long-range direction, while temporal
differences capture local movement. All statistics use the same captured hidden
states, require no extra model call, and produce a fixed-size output for answers
of different lengths.

\actmap{} tests whether a temporal statistic can carry different information at
different depths and coordinates. The representation therefore keeps the layer
and pooled-coordinate axes. In the grouped ablations, removing one channel
family changes AUROC only slightly, whereas collapsing the depth-resolved map
causes a much larger loss. This result supports the full structured map but does
not show that any single channel family is essential.

The same per-example standardization is applied to every retained channel. It
removes absolute scale differences between channels and examples but keeps the
relative pattern across layers and pooled coordinates. The global-normalization
control in the main paper tests whether this pattern adds information beyond
the overall activation scale.

\subsection{Formal Representation and Computational Properties}

For a trajectory $H\in\mathbb{R}^{L\times T\times D}$, let
$P_D:\mathbb{R}^{D}\rightarrow\mathbb{R}^{128}$ denote contiguous adaptive
average pooling over hidden coordinates. Applying it independently to every
$(\ell,t)$ gives
\begin{equation}
  X_{\ell,t,:}=P_D(H_{\ell,t,:}),
  \qquad X\in\mathbb{R}^{L\times T\times128}.
\end{equation}
For channel $c$, let $\phi_c$ be one of the twelve scalar trajectory
statistics listed above. The unpooled channel maps are
\begin{equation}
  S_c[\ell,d]=\phi_c\!\left(X_{\ell,0:T-1,d}\right),
  \qquad S\in\mathbb{R}^{12\times L\times128}.
\end{equation}
Let $A_L$ be ordered adaptive average pooling from $L$ rows to 32 rows, and
let $\mathcal{N}$ denote independent per-channel spatial standardization.
The stored ActMap is therefore the deterministic operator
\begin{equation}
  M=\Phi(H)=\mathcal{N}\!\left(A_L(S)\right)
  \in\mathbb{R}^{12\times32\times128}.
\end{equation}
The learned detector is a separate function
\begin{equation}
  \hat p= f_\theta\!\left(\Phi(H)\right),
  \qquad \hat p\in[0,1],
\end{equation}
trained to estimate the correctness label for a fixed generator and task.
$\Phi$ defines the reusable representation. We do not assume that $f_\theta$ or
its decision boundary transfers without target-domain supervision.

The output shape is independent of $L$, $T$, and $D$ for a decoder-only
transformer whose hidden states can be captured. The map keeps layer and
coordinate order, so the classifier can use their relative structure. During
generation, the implementation reduces each hidden state to 128 pooled
coordinates. It applies the twelve statistics after the pooled sequence is
complete. The stored map has $12\cdot32\cdot128$ values regardless of output
length, instead of $O(LTD)$ full hidden-state values. Because segment boundaries
depend on the final token count, the current implementation temporarily keeps
the pooled $O(LT\cdot128)$ sequence until generation ends. The ablations test
whether this compression retains enough information for correctness prediction.

\section{Primary Classifier}

The main experiments use the compact transformer in
Table~\ref{tab:classifier}. A convolution whose stride equals its kernel size
forms non-overlapping learned patches. It turns the $32\times128$ map into an
$8\times8$ grid of 64 tokens. A learned classification token is prepended.
Learned row and column embeddings preserve both spatial axes without a full
64-position table.

\begin{table*}[t]
\centering
\caption{Exact classifier architecture. The total number of trainable
parameters is 2,401,985.}
\label{tab:classifier}
\small
\setlength{\tabcolsep}{4pt}
\begin{tabular}{lll}
\toprule
Component & Configuration & Output \\
\midrule
Input & 12 channels & $12\times32\times128$ \\
Patch projection & Conv2d, $4\times16$, stride $4\times16$ & $64\times192$ \\
Position & class + factorized $8$ row/$8$ column & $65\times192$ \\
Encoder & 6 pre-norm blocks & $65\times192$ \\
Attention & 6 heads, head width 32 & $65\times192$ \\
Block MLP & $192\rightarrow576\rightarrow192$, GELU & $65\times192$ \\
Readout & LN, $192\rightarrow128\rightarrow1$, GELU & scalar logit \\
\bottomrule
\end{tabular}
\end{table*}

Attention dropout is 0.1. MLP and readout dropout are 0.3, while positional
dropout is 0.15. Stochastic-depth probability increases linearly from 0 in the
first block to 0.05 in the sixth. The patch-projection weights use Xavier
uniform initialization. The class token and factorized positional embeddings
use a truncated normal distribution with standard deviation 0.02; remaining
linear and normalization layers use PyTorch defaults.

The parameter accounting is: 147,648 for patch projection, 3,456 for class and
position parameters, 2,225,664 across the six encoder blocks, 384 for the final
layer normalization, and 24,833 for the readout head. The sigmoid of the scalar
logit is the reported estimated correctness probability.

\section{Optimization and Model Selection}

Each (generator, dataset) detector is trained independently on its balanced
training split with the hyperparameters in Table~\ref{tab:training}.
Validation and test splits are also balanced and disjoint by source key. We use
binary cross-entropy with logits. The positive-class weight is $N_-/N_+$, which
equals 1.0 for every balanced training split in the main experiments.

\begin{table}[t]
\centering
\caption{Training hyperparameters used by every main-table \actmap{} detector.}
\label{tab:training}
\small
\setlength{\tabcolsep}{5pt}
\begin{tabular}{ll}
\toprule
Hyperparameter & Value \\
\midrule
Optimizer & AdamW \\
Initial learning rate & $10^{-3}$ \\
Weight decay & 0.05 \\
Batch size & 64 \\
Maximum epochs & 80 \\
Warm-up & 5 epochs, linear \\
Post-warm-up schedule & cosine, final multiplier 0.01 \\
Early-stopping patience & 20 epochs \\
Selection metric & validation AUROC \\
Gradient-norm clipping & 1.0 \\
Gaussian input noise & $\sigma=0.08$ \\
Mixup & $\alpha=0.2$ \\
Seeds & 42, 123, 456 \\
\bottomrule
\end{tabular}
\end{table}

Gaussian noise is sampled independently for every training map before batching.
For each \actmap{} training batch, a single
$\lambda\sim\operatorname{Beta}(0.2,0.2)$ mixes maps with a random permutation;
the loss is the corresponding convex combination of the two binary losses.
Validation and test maps receive no augmentation. The checkpoint is replaced
whenever validation AUROC strictly improves; ties retain the earlier epoch.
Training stops after 20 epochs without improvement. The selected checkpoint is
then evaluated exactly once on the test split. Main-table values are arithmetic
means over the three seeds; ECE uses ten equal-width probability bins.

The learning-rate multiplier at zero-indexed epoch $e$ is
\begin{equation}
  \eta(e)=
  \begin{cases}
    (e+1)/5, & e<5,\\
    0.01 + 0.495\left[1+\cos\left(\pi\frac{e-5}{80-5}\right)\right], & e\geq5.
  \end{cases}
\end{equation}

\subsection{ACT-ViT Reproduction Protocol}

We evaluate ACT-ViT using the authors' released architecture. Each generated
trajectory is converted to the published
$L_{\mathrm{eff}}=8$ by $N_{\mathrm{eff}}=100$ dense activation tensor. As in
the released preprocessing, the output-token axis is sliced or zero-padded to
the fixed $N_{\max}=100$ positions, and the zero-padded layer axis is max-pooled
to eight groups. We retain the released model-specific linear-adapter branch
and zero-padding rule. We
search the full 24-configuration grid used by the authors: hidden dimensions
$\{128,1024\}$, transformer depths $\{1,3\}$, weight decays
$\{1,10^{-3}\}$, and patch sizes $\{(1,1),(8,1),(4,2)\}$. All configurations
use four attention heads, dropout 0.3, learning rate $10^{-3}$, 15 epochs, and
the published batch size 128. Optimization uses AdamW and binary cross-entropy,
with a cosine schedule and a 10\% step warm-up. The released patience of 30
exceeds the 15-epoch sweep horizon, so every configuration runs for all 15
epochs.

We select the configuration by validation AUROC for seed 42, retrain that
configuration with seeds 123 and 456, and report the three-seed mean. For each
seed, the checkpoint is replaced whenever validation AUROC strictly improves;
ties retain the earlier epoch. We evaluate the selected checkpoint once on the
test split. This matches the authors' test metric at the best-validation epoch;
the test split is not consulted during training or model selection. ACT-ViT
receives the same complete balanced training, validation, and test rows as the other supervised
methods; we do not apply the upstream 10{,}000-example preprocessing cap.
Generators, saved responses, correctness labels, supervision, splits, and
random seeds are fixed across methods. Only the detector and its published
model-selection protocol differ.

\paragraph{Measured cost.} Under the scoring protocol of the main paper
(batch 256, 300 timed repeats, median per-item latency, single NVIDIA L40S),
the validation-selected TriviaQA $\times$ Qwen3-8B configuration
(hidden dimension 128, depth 1, weight decay $10^{-3}$, patch size
$4\times2$; 5.16M parameters) scores at .066\,ms per
answer with a peak of 4{,}949\,MiB of CUDA memory for one batch; the
\actmap{} classifier (2.40M parameters) measured identically scores at
.024\,ms with a 197\,MiB peak. One packed ACT-ViT input tensor
($8 \times 100 \times 4096$, float16) is 6.25\,MiB against \actmap{}'s
96\,KiB map. Training the selected configuration for all three seeds took
1.4--73.7 GPU-minutes per short-form pair and 157.8--195.9 GPU-minutes per
CNN/DailyMail pair on one L40S, excluding the 24-configuration sweep that
selects it. The corresponding three-seed \actmap{} training takes about
54--57 GPU-minutes per TriviaQA pair and about 63 GPU-minutes per
CNN/DailyMail pair; unlike ACT-ViT, it uses one fixed configuration and no
architecture sweep.

\subsection{Transfer Scope and Trainable Adapters}

Transfer freezes the detectors and uses no target labels or adapter updates.
This is stricter than in-domain training. A shared input shape does not guarantee
the same correctness boundary across tasks or generators. The in-domain
ACT-ViT comparison trains its published model-specific adapter; transfer freezes
that adapter. Its near-chance results measure zero-shot reuse. Lightweight
target adaptation may improve them, but
remains outside this study.

\begin{table}[t]
\centering
\caption{In-domain \actmap{} results for Qwen3-32B on balanced test splits (mean over three seeds).}
\label{tab:qwen32}
\small
\begin{tabular}{lccc}
\toprule
Dataset & AUROC & AUPRC & ECE \\
\midrule
TriviaQA & .893 & .892 & .056 \\
NQ-Open & .841 & .816 & .057 \\
GSM8K & .870 & .873 & .082 \\
\bottomrule
\end{tabular}
\end{table}

\section{Reproduction Environment and Retained Artifacts}

The experiments used Python 3.12 with package versions pinned in a lockfile.
Each run manifest records the operating system, CUDA and PyTorch versions,
visible accelerator, configuration, and input-artifact identifiers. We ran all
experiments on NVIDIA L40S GPUs with 48\,GiB of device memory. Experiments with
7--8B models used one L40S; Qwen3-32B experiments used two. The cloud VMs used
x86-64 Ubuntu Noble images. CPU model and host memory varied across generation
and training jobs.

\subsection{Determinism}

The reproduction runner fixes the relevant random seeds and uses deterministic
PyTorch settings. These controls support repeated runs on the recorded software
and hardware stack; exact bitwise agreement across stacks is not claimed.

Before test aggregation, each seed retains the full run configuration, the
classifier constructor arguments, the validation-selected model state, the
per-epoch loss and validation history, and the aggregate test metrics with one
score and probability per balanced test row. The configuration includes split
counts, optimizer, augmentation, determinism settings, and source-artifact
metadata. The three checkpoints for each (generator, dataset) pair are the
frozen detectors used in the transfer experiments.

\section{Classifier Controls}

The classifier ablation keeps the maps and data splits fixed. Logistic
regression maps all 49,152 entries to one logit (49,153 parameters). The matched
MLP flattens the same map, applies one GELU hidden layer and dropout, and emits
one logit. Its hidden width matches the primary classifier's parameter budget.
The $4\times$ control uses four times that budget. Every architecture receives
the same input information, so the comparison tests classifier capacity and
architecture without changing the representation.

\section{Measured Per-Answer Latencies}

\begin{figure}[t]
\centering
\includegraphics[width=\columnwidth]{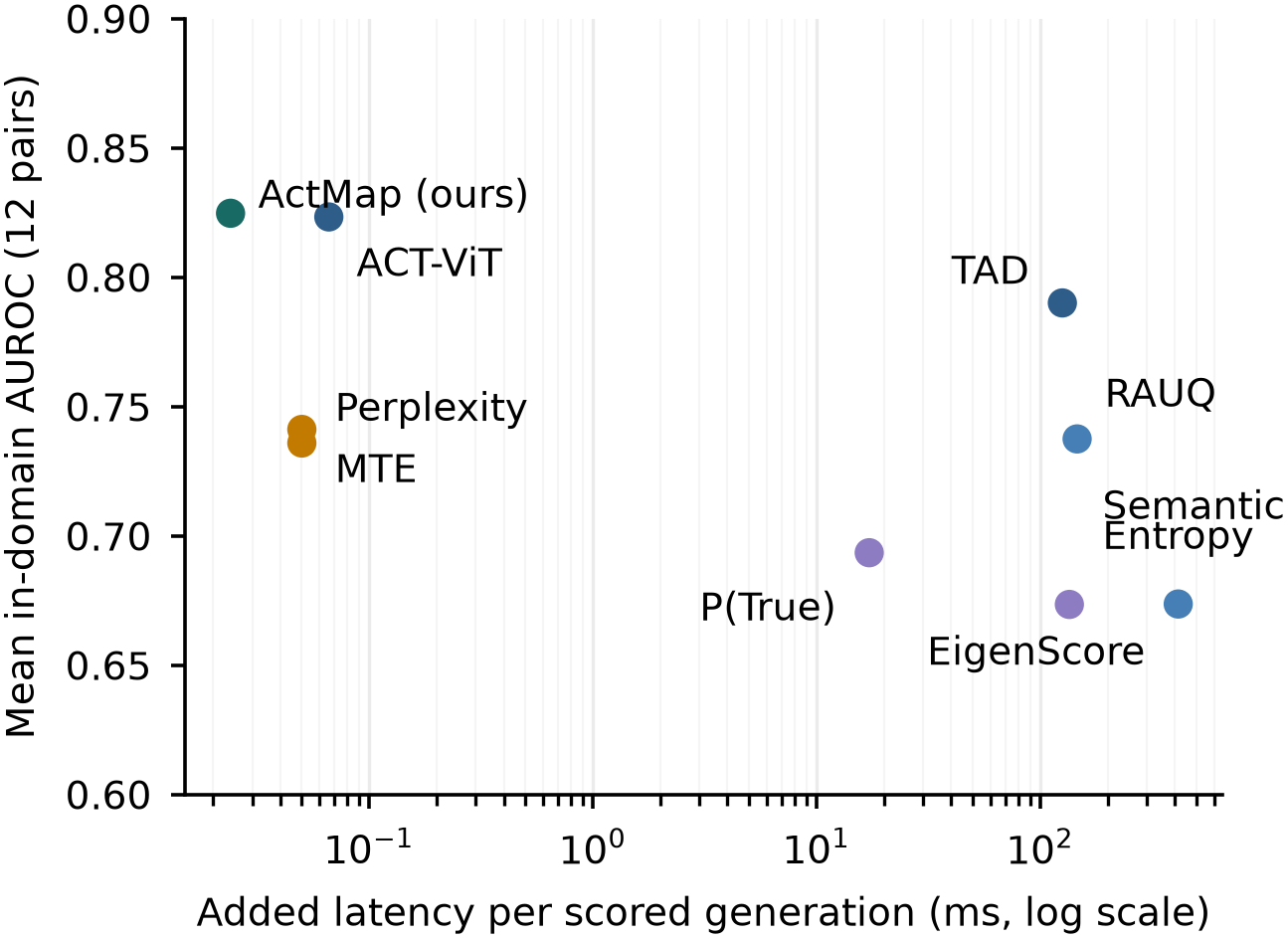}
\caption{Added per-answer latency (log scale; our stack, Table~\ref{tab:latency})
against mean in-domain AUROC over the twelve 7--8B model--dataset pairs of the
main paper. Latencies are implementation- and hardware-specific.}
\label{fig:cost-quality}
\end{figure}

Table~\ref{tab:latency} reports latency after the scored answer has been produced
on our stack (vLLM, one NVIDIA L40S, Qwen3-8B $\times$ TriviaQA, 32 new tokens).
The values can change with the serving stack, kernels, batch size, and output
length.
For Semantic Entropy and EigenScore, sampling cost is the measured 8.5\,ms
marginal decode time times 10 samples. NLI cost increases with output length;
parallel sampling can reduce wall-clock time by using more memory. Figure~
\ref{fig:cost-quality} relates these measurements to mean ranking quality.

\begin{table}[t]
\centering
\caption{Measured per-answer latency beyond producing the answer, on our
implementation and hardware stack.}
\label{tab:latency}
\small
\setlength{\tabcolsep}{4pt}
\begin{tabular}{ll}
\toprule
Method & Added latency \\
\midrule
Perplexity / MTE & negligible post-processing \\
P(True) & 17.2\,ms \\
Semantic Entropy & 85\,ms + 50--291\,ms NLI \\
TAD & 125.6\,ms \\
RAUQ & 146.2\,ms \\
EigenScore & 85\,ms + 329\,ms cov. \\
ACT-ViT & .066\,ms score \\
\actmap{} & no measurable capture; .024\,ms score \\
\bottomrule
\end{tabular}
\end{table}

\section{TAD Latency Protocol}

The TAD benchmark restores the fitted official two-stage checkpoint for
Qwen3-8B on TriviaQA. We time the evaluation path used in the paper:
teacher-forced BF16 inference with eager attention, extraction of top-10
all-layer/all-head attention and token probabilities, and the fitted Ridge
readout. Loading and training are excluded. After eight warm-up rows, we score
128 balanced test rows twice on each of two L40S GPUs. Each GPU runs an isolated
single-GPU workload. The two means are 126.25 and 124.86\,ms, or 125.6\,ms
overall. Across 512 scores, feature extraction averages 110.6\,ms and readout
14.9\,ms. Total latency has a median of 120.8\,ms and a 95th percentile of
158.9\,ms. The experiment archive retains the raw timings and software versions.

\end{document}